\documentclass{article}
\usepackage{spconf,graphicx}
\usepackage{amsfonts}
\usepackage{pifont}
\usepackage{amsmath,amssymb, amsfonts, bm} % define this before the line numbering.
\usepackage{color} 
\usepackage{tabularx, float, booktabs, adjustbox}
\usepackage[hidelinks]{hyperref}
\usepackage{color,soul, graphicx}
\usepackage{mathtools}

\DeclarePairedDelimiter\floor{\lfloor}{\rfloor}
\newcolumntype{Y}{>{\centering\arraybackslash}X}
\usepackage{algorithm}% http://ctan.org/pkg/algorithms
\usepackage[noend]{algpseudocode}%
\usepackage{tikz, amsmath, amssymb, tkz-euclide,tikz-layers}
\usetikzlibrary{intersections,calc,positioning,arrows,arrows.meta}

\title{A Single Graph Convolution is All You Need: Efficient Grayscale Image Classification}
\name{Jacob Fein-Ashley$^{\dagger}$, Sachini Wickramasinghe$^{\dagger}$, Bingyi Zhang$^{\dagger}$, Rajgopal Kannan$^{*}$, Viktor Prasanna$^{\dagger}$}
\address{$^{\dagger}$ University of Southern California, $^{*}$DEVCOM Army Research Office }

\begin{document}
%\ninept
%
\maketitle
\begin{abstract}
Image classifiers for domain-specific tasks like Synthetic Aperture Radar Automatic Target Recognition (SAR ATR) and chest X-ray classification often rely on convolutional neural networks (CNNs). These networks, while powerful, experience high latency due to the number of operations they perform, which can be problematic in real-time applications. Many image classification models are designed to work with both RGB and grayscale datasets, but classifiers that operate solely on grayscale images are less common. Grayscale image classification has critical applications in fields such as medical imaging and SAR ATR. In response, we present a novel grayscale image classification approach using a vectorized view of images. By leveraging the lightweight nature of Multi-Layer Perceptrons (MLPs), we treat images as vectors, simplifying the problem to grayscale image classification. Our approach incorporates a single graph convolutional layer in a batch-wise manner, enhancing accuracy and reducing performance variance. Additionally, we develop a customized accelerator on FPGA for our model, incorporating several optimizations to improve performance. Experimental results on benchmark grayscale image datasets demonstrate the effectiveness of our approach, achieving significantly lower latency (up to $16\times$ less on MSTAR) and competitive or superior performance compared to state-of-the-art models for SAR ATR and medical image classification.
  
 % {\color{red}[to sachini] one sentence briefly describe what we achieve on FPGA.}{\color{blue} We implement our model on state-of-the-art FPGA board -- Xilinx xxx xxx . Compared with state-of-the-art GPU implementation, our FPGA implementation achieves xxx speedup in terms of latency and xxx speedup in terms of throughput. }
\end{abstract}
\section{Introduction}
\label{sec:intro}
As the demand and popularity of real-time systems increase, low-latency machine learning has become increasingly important. With more and more consumers interacting with machine learning models through the cloud, the speed at which those models can deliver results is critical. Consumers expect fast and accurate results; any latency can lead to a poor user experience. Moreover, low-latency machine learning is essential in real-time applications, such as autonomous vehicles or stock market trading, where decisions must be made quickly and accurately. In these scenarios, delays caused by high latency can result in severe consequences and even cause inaccurate downstream calculations~\cite{lowlatency}. 

 A particular instance where low-latency machine learning is needed is grayscale image classification for SAR ATR. For example, a targeting system on a satellite is costly, and decisions must be made using SAR efficiently and accurately. Examples like this are where low-latency grayscale image classification comes into play. It is often the case that image classifiers work on RGB datasets and grayscale image datasets, but seldom do modern image classifiers focus solely on the grayscale setting. RGB models are overkill for the grayscale setting, as the grayscale problem allows us to focus on a single channel.  Models focusing on grayscale image classification are naturally more efficient, as they can concentrate on a single channel rather than three. Thus, many image classifiers that generalize to the grayscale image classification are not truly optimized for the grayscale case. For these reasons, we present a lightweight grayscale image classifier capable of achieving up to $16 \times$ lower latency than other state-of-the-art machine learning models on the MSTAR dataset.

From a trustworthy visual data processing perspective, the demand for grayscale image classification requires data to be collected from various domains with high resolution and correctness so that we can train a robust machine learning model. Additionally, recent advancements in machine learning rely on convolutional neural networks, which often suffer from high computation costs, large memory requirements, and many computations needed, resulting in poor inference latency, poor scalability, and weak trustworthiness. 

The inherent novelties of our model are as follows: Our proposed method is the first to vectorize an image in a fully connected manner and input the resultant into a single-layer graph convolutional network (GCN). We also find that a single GCN layer is enough to stabilize the performance of our shallow model. Additionally, our proposed method benefits from a batch-wise attention term, allowing our shallow model to capture interdependencies between images and form connections for classification. Finally, by focusing on grayscale imagery, we can focus on a streamlined method for grayscale image classification rather than concentrating on the RGB setting. A result of these novelties is extremely low latency and high throughput for SAR ATR and medical image classification.

% With the recent technological advances, modern FPGAs contain many hardware resources~\cite{fpgaperformance}, including DSPs, LUTs, BRAMs, and URAMs. The programmable nature of FPGAs allows users to develop a customized data path and on-chip memory organization, leading to high-performance implementations. Consequently, FPGAs have emerged as an appealing option for executing time-sensitive machine learning tasks with reduced latency and power. For instance, FPGAs have been used for accelerating machine learning~\cite{fpgaAccel} and graph analytic tasks~\cite{graphAgile}. Given that our model is a lightweight image classifier, we find it suitable to be deployed on an FPGA. We can perform inference on the FPGA without returning the intermediate results to external memory. This ensures low-latency inference by capitalizing the fine-grained data parallelism inherent in FPGAs. In contrast, CPUs and GPUs exploit coarse-grained thread-level parallelism and have complex cache hierarchies, which are unsuitable for low-latency inference. Thus, this  paper makes the following contributions:

\begin{itemize}
    \item We present a lightweight, graph-based neural network for grayscale image classification. Specifically, we (1) apply image vectorization, (2) construct a graph for each batch of images and apply a single graph convolution, and (3) propose a weighted-sum mechanism to capture batch-wise dependencies. 

    \item We implement our proposed method on FPGA, including the following design methodology: (1) a portable and parameterized hardware template using high-level synthesis, (2) layer-by-layer design to maximize runtime hardware resource utilization, and (3) a one-time data load strategy to reduce external memory accesses.
    
    \item  Experiments show that our model achieves competitive or leading accuracy with respect to other popular state-of-the-art models while vastly reducing latency and model complexity for SAR ATR and medical image classification.

    \item We implement our model on a state-of-the-art FPGA board, Xilinx Alveo U200. Compared with state-of-the-art GPU implementation, our FPGA implementation achieves comparable latency and throughput with state-of-the-art GPU, with only $1/41$ of the peak performance and $1/10$ of the memory bandwidth.  
        
\end{itemize}

\section{Problem Definition}
\label{sec:problemdefinition}
The problem is to design a lightweight system capable of handling high volumes of data with low latency. The solution should be optimized for performance and scalability while minimizing resource utilization, a necessary component of many real-time machine learning applications. The system should be able to process and respond to requests quickly, with minimal delays. High throughput and low latency are critical requirements for this system, which must handle many concurrent requests without compromising performance. We define latency and throughput in the following ways:

\begin{align*}
    \text{Throughput} &= \frac{\text{Total number of images processed}}{\text{Total inference time}} \\ \\
    \text{Latency} &= \text{Total time for a single inference}
\end{align*}

Latency refers to the total time (from start to finish) it takes to gather predictions for a model in one batch. A lightweight machine learning model aims to maximize throughput and accuracy while minimizing latency.

\section{Related Work}
\label{sec:relatedwork}
\subsection{MLP Approaches}
Our model combines various components of simple models and is inherently different from current works in low-latency image classification. Some recent architectures involve simple MLP-based models. Touvron et al. introduced ResMLP~\cite{resmlp}, an image classifier based solely on MLPs. ResMLP is trained in a self-supervised manner with patches interacting together. Touvron et al. highlight their model's high throughput properties and accuracy. ResMLP uses patches from the image and alternates linear layers where patches interact and a two-layer feed-forward network where channels interact independently per patch.
Additionally, MLP-Mixer~\cite{mlpmixer} uses a similar patching method, which also attains competitive accuracy on RGB image datasets compared to other CNNs and transformer models.  Our proposed method uses the results from a single-layer MLP to feed into a graph neural network, during which we skip the information from the three-channel RGB setting and only consider the single-channel grayscale problem. This is inherently different from the methods mentioned earlier, as they use patching approaches while we focus on the vectorization of pixels.

\subsection{Graph Image Construction Methods}
The dense graph mapping that utilizes each pixel as a node in a graph is used and mentioned by~\cite{sanchez-lengeling2021a, goyal2022graph}. For this paper, we employ the same terminology. Additionally, Zhang et al. presented a novel graph neural network architecture and examined its low-latency properties on the MSTAR dataset using the dense graph~\cite{Zhang_2022}. Our proposed method differs from dense graph methods, as we vectorize an image rather than using the entire grid as a graph.

Han et al.~\cite{han2022vision} form a graph from the image by splitting the image into patches, much like a transformer. A deep graph neural network learns on the patches similarly to a transformer but in a graph structure. Our structure does not form a graph where each patch is a node in a graph; instead, we create a graph from the resultant of a vectorized image passed through a fully connected layer.

Mondal et al. proposed applying graph neural networks on a minibatch of images~\cite{minibatch}. Mondal et al. claim that this method improves performance for adversarial attacks. We use the proposed method to stabilize the performance of a highly shallow model. The graph neural network, in this case, allows learning to be conducted in a graph form, connecting images containing similar qualities.

Besides the model proposed by Zhang et al., all the methods mentioned focus on the RGB setting. This is overkill for grayscale image classification. Focusing on a single channel allows us to develop a more streamlined solution rather than forcing a model to operate on RGB datasets and having the grayscale setting come as an afterthought. Doing so allows us to reduce computational costs.

\section{Overview and Architecture}
\label{sec:overview}
\begin{figure}[ht]
\centering
\includegraphics[scale=0.215]{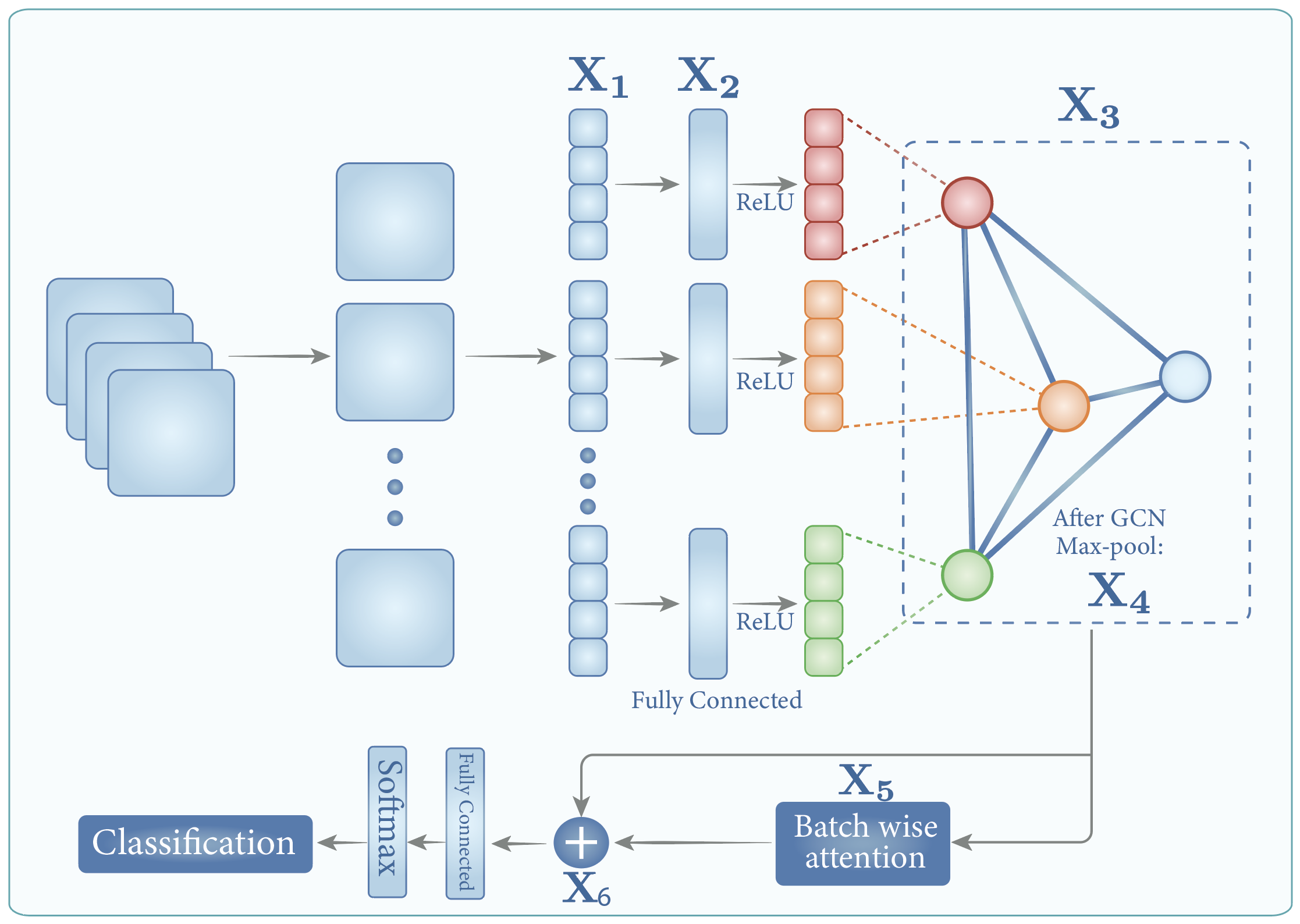}
\caption{GECCO Architecture}
\label{fig:gecco}
\end{figure}

This section describes our model architecture (GECCO: Grayscale Efficient Channelwise Classification Operative). The overall process is summarized in Figure~\ref{fig:gecco}.

\textbf{Overall Architecture. }Many existing methods do not focus on the latency of their design and its implications. Additionally, the vast majority of image classification models focus on the performance of their work in the RGB setting, rarely citing the performance of datasets in various domains. We address these problems by presenting a novel architecture focused on low latency and the grayscale image setting. 

Our model vectorizes a batch of images, allowing us to use a fully connected layer (FC) pixel-wise for low computation time rather than relying on convolutional neural networks. We vectorize the input images and input them into a fully connected layer. Then, we use a graph convolutional layer to learn similarities between images batch-wise. We then apply a batch-wise attention term, which is inputted into an FC for classification.\footnote{We make our code publicly available at \url{https://github.com/GECCOProject/GECCO}}

\textbf{Image Vectorization. }
For each image in a batch, we view the image as a vector. For a tensor $\mathbf{X}\in \mathbb{R}^{B \times H \times W}$ where $B$ is the batch size, 
$H$ and $W$ are the height and width of an image; we flatten the tensor to $\mathbf{X}_1\in\mathbb{R}^{B \times (H \cdot W)}$. 
Viewing an image as a vector allows our model to skip the traditional convolutional neural network, which views the image as a grid and cuts computation time.

\textbf{Fully Connected Layer. }
We input $\mathbf{X}_1$ into an FC layer with output dimensionality $D_{out}$.
Formally, $\mathbf{X}_2 = \sigma( \mathbf{X}_1\mathbf{W}_1 + \mathbf{b}_1 )$, where $\sigma$ is the ReLU function, 
$\mathbf{W}_1$ is a learned weight matrix, $\mathbf{b}_1$ is a bias term, and $\mathbf{X}_2 \in \mathbb{R}^{B \times D_{out}}$.

After the fully connected layer, we apply a dropout layer and the ReLU function to $\mathbf{X}_2$, yielding $\mathbf{X}_3$,
such that the resultant dimensionality of $\mathbf{X}_3$ is $\mathbb{R}^{B \times D_{out}}$.

\textbf{Graph Construction. }
We construct a graph batch-wise from $\mathbf{X}_3$. This means that for each batch, a vectorized image is each node in the graph with feature size 
$\mathbb{R}^{D_{out}}$, and each image is connected to every other image in a batch. Formally, we calculate the adjacency matrix 
$\mathbf{A}$ as $\mathbf{A}_{ij} = 1$, which connects all nodes. 

\textbf{Graph Convolution. }
Our single graph convolutional layer learns from similar features of images within its minibatch.
Generally, a graph convolutional layer updates the representations of nodes by aggregating each node's neighbor's representation.
We can write a graph convolutional layer as $\bm{h^{\prime}}_i = f_{\theta} \left( \bm{h}_i, \text{\scriptsize{AGGREGATE}}\left(\{\bm{h_j} \mid j \in \mathcal{N}_i \}\right) \right)$.
In our case, the input for each node $\bm{h_i}$ is the output from each vector in $\mathbf{X}_3$.

Applying graph convolution to $\mathbf{X}_3$ yields $\mathbf{X}_4$.
% For graph convolution, we need twhe equation for X4 in terms of W, A, and X3.
Formally, $\mathbf{X}_4 = \sigma \left( \mathbf{A}\mathbf{X}_3\mathbf{W}_2 \right)$, where $\mathbf{W}_2$ is a learned weight matrix and $\sigma$ represents the sigmoid function.
After the graph convolution, we apply batch normalization and max-pooling operations to $\mathbf{X}_4$, resulting in a
dimensionality of $\mathbb{R}^{B \times \floor{D_{out} / 2}}$.

\textbf{Batch-wise Attention, Residual Connections, \& Output.}
We propose a batch-wise attention term defined as
\begin{align*}
\mathbf{X}_5 &= \left( \frac{\sigma \left( \mathbf{X}_4\mathbf{X}^{\top}_4 \right)}{\sum_{i=1}^B
\sigma \left( \mathbf{X}_4\mathbf{X}_4^{\top} \right)_{i}} \right) \mathbf{X}_4
\end{align*}
where $\sigma$ is the sigmoid function. This term allows the model to capture similar features from each image to another batch-wise.
 
The residual connection is defined $\mathbf{X}_6 = \mathbf{X}_5 +\mathbf{X}_4$. The residual term makes the learning process easier and more stable.
By multiplying a softmax-like term with the output of the previous graph convolution, 
we weigh the correspondence of each image compared to other similar images within similar images batch-wise. We then feed the residual term into an FC inputted into the softmax function for classification results.

\subsection*{Model Structure Discussion}
We justify our model's design choices by considering the following theoretical aspects.
\begin{enumerate}
    \item The batch-wise attention term allows the model to further capture similar features from each image to another batch-wise. Relating similar properties from images to each other boosts accuracy in our case
    of a shallow model. Additionally, our batch-wise attention term is similar in spirit to the mechanism proposed by~\cite{cheng2021ba2m}, which allows the model to capture long-range dependencies across the entire image.
    \item The \emph{batch-size} hyperparameter is crucial in our model. 
    A larger batch size allows the model to capture more dependencies across images, which is crucial for understanding complex image patterns.
    We refer to the work of~\cite{hu2021on} for a detailed analysis of the impact of batch size on the performance of GNNs.
    \item If the batch size for a given dataset is $1$, the model eliminates the graph construction phase, making the term $\mathbf{X}_3$ fed directly into the FC and softmax for classification.
    \item The residual connection term makes the learning process easier and more stable.
    we refer to~\cite{yu2019identity} for a more detailed analysis of the impact of residual connections on shallow models.
\end{enumerate}

\section{Experiments}
\label{sec:experiments}
\subsection{Datasets}
% Datasets from several domains are examined to gauge the effectiveness of GECCO in diverse settings. We use the popular MNIST dataset~\cite{mnist}, Fashion-MNIST~\cite{fashionmnist}, MSTAR, and CXR~\cite{chest} summarized in the following manner:

Datasets from several domains are examined to gauge the effectiveness of GECCO in diverse settings. We use the SAR ATR dataset, MSTAR, and a medical imaging dataset CXR~\cite{chest}.

\begin{itemize}

% \item MNIST is a grayscale handwritten dataset with $(28, 28)$ pixel image sizes and $10$ different object categories. The training size for this dataset is $60000$, and the testing size is $10000$.

% \item Fashion-MNIST contains $(28, 28)$ sized grayscale images from $10$ categories with a training size of $60000$, and a testing size of $10000$.

\item MSTAR is a  SAR ATR dataset with a training size of $2747$ and testing size of $2425$ SAR images of $10$ different vehicle categories. We resize each image in the dataset to $(128, 128)$ pixels.

\item CXR is a chest X-ray dataset containing 5863 X-ray images and 2 categories (Pneumonia/Normal). The images are $(224, 224)$ pixels. The training size is $5216$, and the testing size is $624$.

\end{itemize}

Our goal is to create a real-time system. That is, we wish to minimize the inference latency and maximize the throughput of our model while maintaining leading or competitive accuracy on its respective dataset. In the following sections, we measure the inference latency and throughput, as described in section~\ref{sec:problemdefinition}.

\subsection{Results}
\subsubsection{Backbone}
\label{sec:backbone}
For Table~\ref{convlayers}, we choose ResConv as the backbone of our model because it has the most desirable characteristics for applying a graph convolutional layer.

\begin{table}[H]
 \caption{Performance of GCN Layers on MSTAR}
    \label{convlayers}
    \centering 
    \resizebox{\columnwidth}{!}{
        \begin{tabularx}{1.5\columnwidth}{@{}lYYYYY@{}}
        \toprule
        \textbf{Convolutional Layer} & \textbf{Top-1 Accuracy} & \textbf{Throughput ($\text{imgs}/\text{ms}$)} & \textbf{Latency (ms)}\\
        \midrule
        GCN~\cite{gcnconv} & $98.89\%$ & $50.04 \pm 6.85$ &  $5.86 \pm 0.98$ \\
        TAGConv~\cite{tagconv}  & $99.05\%$ &  $47.87 \pm 7.11$ &  $6.24 \pm 1.32$ \\ 
        SAGEConv~\cite{sageconv} & $99.08\%$ & $51.77 \pm 8.19$ &  $5.95 \pm 0.87$ \\
        ChebConv~\cite{chebconv}  & $98.56\%$ & $45.37 \pm 5.99$ &  $6.83 \pm 1.27$ \\
        ResConv~\cite{bresson2018residual} & $\bf{99.29\%}$ &  $\bf{52.98 \pm 9.04}$ &  $\bf{5.22 \pm 1.03}$\\
        \bottomrule
        \end{tabularx}
        }
\end{table}

We use the following hyperparameters listed in Table \ref{featurelength} for our experiments.
\begin{table}[H]
    \caption{Feature Lengths, Optimizer, and Batch Size for Each Dataset}
    \label{featurelength}
    \centering
    \resizebox{\columnwidth}{!}{
        \begin{tabularx}{1.5\columnwidth}{@{}lYYYYY@{}}
        \toprule
        \textbf{Dataset} & \textbf{Feature Length} & \textbf{Optimizer} & \textbf{Batch Size} \\
        \midrule
        MSTAR & $86$ & Adam & $64$ \\
        CXR & $112$ & Adam & $64$ \\
        \bottomrule
        \end{tabularx}
        }
\end{table}

\vspace{-2em}
\subsubsection{Experimental Performance}
Experimental performance includes the top-1 accuracy, inference throughput, and inference latency. We perform our inference batch-wise as a means to reduce latency. These metrics vary across each dataset.

% In Tables~\ref{mnistp}, \ref{fashionmnistp}, \ref{mstarp}, and \ref{cxr}, we present a summary of our findings. We report the best-performing accuracy, average throughputs, and latencies with their standard deviations. Our model outperforms every other model in terms of throughput and latency across all datasets, leads accuracy on the MSTAR dataset, and performs competitively in terms of accuracy on all datasets. 

We summarize our findings in Tables~\ref{mstarp} and \ref{cxr}. We report the best-performing accuracy, average throughputs, and latencies with their standard deviations. Our model outperforms every other model in terms of throughput and latency across all datasets, leads accuracy on the MSTAR dataset, and performs competitively in terms of accuracy on all datasets. 

We perform the remaining experiments on a state-of-the-art NVIDIA RTX A5000 GPU. Additionally, we compare our model to the top-performing variants of VGG~\cite{vgg}, the variant of the popular ViT~\cite{vit}, the ViT for small-sized datasets (SS-ViT)~\cite{smallvit}, FastViT~\cite{fastvit}, Swin Transformer~\cite{swin}, and ResNet~\cite{resnet} models. We use the open-source packages PyTorch and HuggingFace for model building and the PyTorch Op-Counter for operation counting. Performing the remaining experiments on the same hardware system is vital in fostering a fair comparison for each model.

\vspace{-1em}
% \begin{table}[H]
%     \caption{MSTAR Performance}
%     \label{mstarp}
%     \centering 
%     \resizebox{\columnwidth}{!}{
%         \begin{tabularx}{1.35\columnwidth}{@{}lYYYYY@{}}
%         \toprule
%         \textbf{Model} & \textbf{Top-1 Accuracy} & \textbf{Throughput ($\text{imgs}/\text{ms}$)} & \textbf{Latency (ms)}\\
%         \midrule
%         Swin-T & $86.04\%$ & $5.44 \pm 0.37 $ &  $46.98 \pm 3.20$ \\
%         SS-ViT & $95.61\%$ & $  9.14 \pm 1.72$ &  $ 27.97 \pm 5.26$ \\
%         VGG16  & $93.13\%$ &  $6.75 \pm 1.34$ &  $37.89 \pm 7.52$ \\ 
%         FastViT & $91.78\%$ & $4.16 \pm 0.52$ & $61.44  \pm 7.69$\\
%         ResNet34 & $98.64\%$ & $12.44 \pm 0.84$ & $20.48 \pm 1.39$\\
%         GECCO & $\bf{99.29\%}$  &  $\bf{52.98 \pm 9.04}$ &  $\bf{5.22 \pm 1.03}$ \\
%         \bottomrule
%         \end{tabularx}
%         }
% \end{table}

\begin{table}[H]
    \caption{MSTAR Performance}
    \label{mstarp}
    \centering 
    \resizebox{\columnwidth}{!}{
        \begin{tabularx}{1.35\columnwidth}{@{}lYYYYY@{}}
        \toprule
        \textbf{Model} & \textbf{Top-1 Accuracy} & \textbf{Throughput ($\text{imgs}/\text{ms}$)} & \textbf{Latency (ms)}\\
        \midrule
        Swin-T & $86.04\%$ & $1.36 \pm 0.10 $ &  $46.98 \pm 3.20$ \\
        SS-ViT & $95.61\%$ & $ 2.29 \pm 0.43$ &  $ 27.97 \pm 5.26$ \\
        VGG16  & $93.13\%$ &  $1.69 \pm 0.33$ &  $37.89 \pm 7.52$ \\ 
        FastViT & $91.78\%$ & $1.04 \pm 0.13$ & $61.44  \pm 7.69$\\
        ResNet34 & $98.64\%$ & $3.13 \pm 0.22$ & $20.48 \pm 1.39$\\
        GECCO & $\bf{99.29\%}$  &  $\bf{12.26 \pm 2.42}$ &  $\bf{5.22 \pm 1.03}$ \\
        \bottomrule
        \end{tabularx}
        }
\end{table}

\vspace{-2em}
\begin{table}[H]
    \caption{CXR Performance}
    \label{cxr}
    \centering 
    \resizebox{\columnwidth}{!}{
        \begin{tabularx}{1.4\columnwidth}{@{}lYYYYY@{}}
        \toprule
        \textbf{Model} & \textbf{Top-1 Accuracy} & \textbf{Throughput ($\text{imgs}/\text{ms}$)} & \textbf{Latency (ms)}\\
        \midrule
        Swin-T & $73.66\%$ & $0.27 \pm 0.05$ &  $236.71 \pm 46.09$ \\
        SS-ViT  & $71.09\%$ & $1.03 \pm 0.21$ &  $62.35 \pm 12.85$ \\
        VGG16   & $\bf{82.01\%}$ & $0.76 \pm 0.25$ &  $84.10 \pm 28.43$ \\
        FastViT & $75.46\%$ & $1.06 \pm 0.14$ &  $60.30 \pm 14.24$ \\
        ResNet34  & $78.31\%$ & $0.60 \pm 0.11$ &  $105.84 \pm 19.39$ \\
        GECCO  & $77.57\%$ & $\bf{2.63 \pm 0.55}$ &  $\bf{24.32 \pm 5.08}$ \\
        \bottomrule
        \end{tabularx}
        }
\end{table}

\subsubsection{Model Complexity Metrics}
Model complexity metrics for this paper include the number of multiply-accumulate operations (MACs), the number of model parameters, the model size, and the number of layers. In other words, suppose accumulator $a$ counts an operation of arbitrary $b,c \in \mathbb{R}$.  We count the number of multiply-accumulate operations as $a \leftarrow a + (b \times c)$. Additionally, the layer count metric is an essential factor of latency. Decreasing the number of layers will also improve the latency of a model's inference time. The goal of an effective machine learning model is to maximize throughput while minimizing the number of MACs and the number of layers, in our case. 

We measure the model complexity of our model against other popular machine learning models that we have chosen in Table~\ref{analytical}. Our model outperforms in all categories regarding our chosen model complexity metrics, highlighting its lightweightness.
\begin{table}[H]
    \caption{Model Complexity Metrics}
    \label{analytical}
    \centering 
    \resizebox{\columnwidth}{!}{
        \begin{tabularx}{1.35\columnwidth}{@{}lYYYYYY@{}}
        \toprule
        \textbf{Model} & \textbf{\# MACs} & \textbf{\# Parameters} & \textbf{Model Size (Mb)} &\textbf{\# Layers}  \\
        \midrule
        Swin-T   & $2.12\times 10^{10}$ & $2.75\times 10^7$ &  $109.9$ &  $167$\\
        SS-ViT  & $1.55 \times 10^{10}$ & $4.85 \times 10^6 $ &  $19.62$ &  $79$\\
        VGG16   & $9.51\times 10^{9}$ & $4.69 \times 10^6 $ &  $18.75$ &  $20$\\
        FastViT & $7.16 \times 10^{8}$ & $4.02 \times 10^{6}$ &  $16.1$ &  $226$\\
        ResNet34  & $4.47\times 10^{9}$ & $2.13 \times 10^{7} $ &  $85.1$ &  $92$\\
        GECCO  & $\bf{5.10 \times 10^{4}}$ & $\bf{5.08 \times 10^4} $ &  $\bf{0.19}$ &  $\bf{16}$\\
        \bottomrule
        \end{tabularx}
        }
\end{table}

\subsubsection{Ablation Study}

We perform an ablation study to verify that the components of our proposed model contribute positively to the overall accuracy of the MSTAR dataset.
\begin{table}[H]
    \caption{Ablation Study}
    \label{ablation}
    \centering 
    \resizebox{\columnwidth}{!}{
        \begin{tabularx}{1.2\columnwidth}{@{}lYY@{}}
        \toprule
        \textbf{Mini-batch GNN}  & \textbf{Weighted Sum Residual Term} & \textbf{Accuracy on MSTAR} \\
        \midrule
        \checkmark & \checkmark & $99.29\%$ \\
        \ding{55} & \checkmark & $97.94\%$  \\
        \checkmark & \ding{55} & $88.04\%$  \\
        \ding{55} & \ding{55} & $78.64\%$  \\
        \bottomrule
        \end{tabularx}
        }

\end{table}

\begin{figure}[ht]
\centering
\includegraphics[scale=0.7]{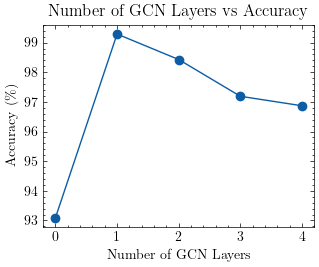}
\caption{Accuracy on MSTAR vs. Number of Graph Convolutional Layers}
\label{fig:gcnlayers}
\end{figure}

Additionally, we find that only a single graph convolutional layer is enough to reduce the variance and increase the accuracy of our model. Refer to Figure~\ref{fig:gcnlayers}.

\vspace{-2em}
\subsection{Discussion}
Across multiple datasets, GECCO achieves leading or competitive accuracy compared to other state-of-the-art image classifiers. GECCO outperforms other machine learning models regarding model complexity, highlighting our model's low latency and lightweight properties.

It is difficult for our model to generalize to the RGB setting. We attribute this challenge to the vectorization process that our model uses. Learning on three channels poses a complexity challenge, as GECCO is very shallow and simple, thus making it challenging to learn on three separate channels. Additionally, our model is optimized for a low-complexity dataset regime, as datasets like CIFAR and ImageNet are much too complex for our shallow model, as they pose a complexity challenge.

Our proposed method does not make use of positional embeddings or class tokens. GECCO can learn essential features using the weighted residual term. Additionally, we tested the addition of positional embeddings and class tokens and found no improvement in accuracy across various datasets. We note that the $\mathbf{X}_5$ attention-like term adds positional awareness to the model.

\subsection{FPGA Implementation}

We develop an accelerator for the proposed model on state-of-the-art FPGA, Xilinx Alveo U200~\cite{alveoU200}, to further highlight the model's efficiency and compatibility with hardware. It has 3 Super Logic Regions (SLRs), 4 DDR memory banks, 1182k Look-up tables, 6840 DSPs, 75.9 Mb of BRAM, and 270 Mb of URAM. The FPGA kernels are developed using the Xilinx High-level Synthesis (HLS) tool to expedite the design process.

\begin{figure}[ht]
\centering
\includegraphics[scale=0.4]{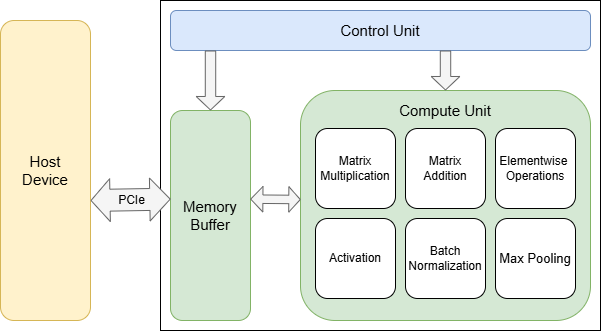}
\caption{Overview of Hardware Architecture}
\label{fig:hw}
\end{figure}

Our FPGA design incorporates several novel features: (1) \emph{Portability of the design}: We design a parameterized hardware template using HLS. It is portable to different FPGA platforms, including embedded and data-center FPGAs. We present our hardware mapping algorithm in Algorithm \ref{alg:Inference-Algorithm}. (2) \emph{Resource sharing}: The model is executed layer-by-layer. Each layer in the model is decomposed into basic kernel functions. The basic kernel functions, including matrix multiplication, elementwise activation, column-wise and row-wise summations, max pooling, and various other elementwise operations, are implemented separately and subsequently invoked within their corresponding layers. Due to the reuse of these fundamental kernel functions across multiple layers, FPGA resources are shared among the different layers, maximizing runtime hardware resource utilization. (3) \emph{Single-load strategy}:  We employ a one-time data load strategy to load the required data from DDR only once. All other data required for the computations are stored in on-chip memory, reducing inference latency. Figure~\ref{fig:hw} illustrates the overall hardware architecture of our design.  

We utilize the Vitis tool~\cite{vitis} for hardware synthesis and place-and-route to determine the achieved frequency. The Vitis Analyzer tool is then used to report resource utilization and the number of clock cycles. The latency is calculated by multiplying the achieved clock period by the number of cycles. Table~\ref{hw} reports the results obtained for the MSTAR dataset. Given the compact design and resource efficiency of the model, it can be accommodated within a single SLR. Hence, we deploy multiple accelerator instances across multiple SLRs, each with one instance. This increases the inference throughput. Table~\ref{hw} shows the latency obtained for a single inference and the throughput achieved by running the design on 3 SLRs concurrently.

\begin{table}[!ht]
\centering
\caption {Comparison with state-of-the-art GPU platform}
\begin{adjustbox}{max width=0.48\textwidth}
\begin{tabular}{c|cc}
 \toprule
           %\\ \midrule
& \textbf{GPU}  & \textbf{Our Design}  \\  \midrule
\midrule 
Platform   & NVIDIA A5000  &  Alveo  U200   \\  

{Technology}    &  Samsung 8 nm  &  TSMC  16 nm  \\ 
{Frequency}   & 1.17 GHz & 200 MHz \\  
Peak Performance (TFLOPS) & 27.7 & 0.66 \\ 
{On-chip Memory} & 6 MB & 35 MB   \\ 
{Memory Bandwidth} & 768 GB/s  & 77 GB/s \\ \midrule
 Latency on MSTAR (ms) & 5.22 & 5.65 \\
Throughput on MSTAR (imgs/ms) & 12.26 & 33.98 \\ \bottomrule
\end{tabular}
\end{adjustbox}
\label{tab:platform-specifications}
\end{table}

\begin{algorithm}
\caption{Hardware Mapping Algorithm (See the definition of layer in Section \ref{sec:overview})}
\label{alg:Inference-Algorithm}
\begin{algorithmic}[1]
\renewcommand{\algorithmicrequire}{\textbf{Input:}}
\renewcommand{\algorithmicensure}{\textbf{Output:}}
 \Require Model $f()$ and the input images;
 \Ensure Execution result;
\For{each layer $i$ in $f()$}
    \If{ layer $i$ is a fully connected layer }
        \State Map to the matrix multiplication unit
    \EndIf
    \If{layer $i$ is a graph convolution layer}
        \State Map to the matrix multiplication unit
        \State Map to the activation unit
        \State Map to the elementwise operation unit
        \State Map to the matrix addition unit
    \EndIf
    \If{layer $i$ is a batch-wise attention}
        \State Map to the matrix multiplication unit
        \State Map to the activation unit
        \State Map to the elementwise operation unit
    \EndIf
        \If{layer $i$ is a max pooling layer}
        \State Map to the max pooling unit
    \EndIf
        \If{layer $i$ is a activation layer}
        \State Map to the activation unit
    \EndIf
        \If{layer $i$ is a batch normalization layer }
        \State Map to the batch normalization unit
    \EndIf
\EndFor
\end{algorithmic}
\end{algorithm}

We compare our FPGA implementation with the baseline GPU implementation. The GPU baseline is executed on an NVIDIA RTX A5000 GPU, which operates at 1170 MHz and has a memory bandwidth of 768 GB/s. However, the FPGA operates at 200 MHz and has an external memory bandwidth of 77 GB/s. We compare the hardware features of the two platforms in Table~\ref{tab:platform-specifications}. Although the GPU has higher peak performance ($41\times$) and memory bandwidth ($10\times$), our FPGA implementation achieves a comparable latency of $5.65~ms$ and an improved throughput of $33.98~imgs/ms$.

% These results are attributed to the optimizations mentioned above that were adopted in our implementation.

\vspace{-2em}

\begin{table}[H]
    \caption{Resource Utilization (per SLR), Latency, and Throughput for MSTAR dataset}
    \label{hw}
    \centering 
    \resizebox{1\columnwidth}{!}{
        \begin{tabularx}{1.2\columnwidth}{@{}l@{}YY}
        \toprule
        % \textbf{~}  & \textbf{~} \\
        % \midrule
        Latency &   $5.65$ \textit{ms} \\
        Throughput  & $33.98$ \textit{imgs/ms} \\
        \midrule
        BRAMs & $956~(22\%)$  \\
       URAMs & $228~(24\%)$ \\
        DSPs  & $1226~(17\%)$  \\
        LUTs  & $459K~(38\%)$  \\
        FFs  & $597K~(25\%)$  \\
        \bottomrule
        \end{tabularx}
        }
\end{table}
\vspace{-2em}

\section{Conclusion and Future Work}
This work introduced a novel architecture combining fully connected and graph convolutional layers, benchmarked on popular grayscale image datasets. The model demonstrated strong performance and low complexity, highlighting the importance of lightweight, low-latency image classifiers for various applications. Its efficacy was shown across SAR ATR and medical image classification, with an FPGA implementation underscoring its hardware friendliness. Key innovations include using a single-layer GCN, which, along with batch-wise attention, enhances accuracy and reduces variance. Future work should explore extending this approach to color image datasets and other domains, optimizing the architecture for even greater efficiency, and further investigating the potential of graph neural networks in shallow models.

\section{Acknowledgement}
This work is supported by the DEVCOM Army Research Lab (ARL)
under grant W911NF2220159. \textbf{Distribution Statement A:} Approved for public release. Distribution is unlimited.

\vfill\pagebreak
% \clearpage

\bibliographystyle{IEEEbib}
\bibliography{refs}

\begin{thebibliography}{10}

\bibitem{lowlatency}
Kaoru Ota, Minh~Son Dao, Vasileios Mezaris, and Francesco G. B.~De Natale,
\newblock ``Deep learning for mobile multimedia: A survey,''
\newblock {\em ACM Trans. Multimedia Comput. Commun. Appl.}, vol. 13, no. 3s,
  jun 2017.

\bibitem{resmlp}
Hugo Touvron, Piotr Bojanowski, Mathilde Caron, Matthieu Cord, Alaaeldin
  El-Nouby, Edouard Grave, Gautier Izacard, Armand Joulin, Gabriel Synnaeve,
  Jakob Verbeek, and Hervé Jégou,
\newblock ``Resmlp: Feedforward networks for image classification with
  data-efficient training,'' 2021.

\bibitem{mlpmixer}
Ilya Tolstikhin, Neil Houlsby, Alexander Kolesnikov, Lucas Beyer, Xiaohua Zhai,
  Thomas Unterthiner, Jessica Yung, Andreas Steiner, Daniel Keysers, Jakob
  Uszkoreit, Mario Lucic, and Alexey Dosovitskiy,
\newblock ``Mlp-mixer: An all-mlp architecture for vision,'' 2021.

\bibitem{sanchez-lengeling2021a}
Benjamin Sanchez-Lengeling, Emily Reif, Adam Pearce, and Alexander~B.
  Wiltschko,
\newblock ``A gentle introduction to graph neural networks,''
\newblock {\em Distill}, 2021,
\newblock https://distill.pub/2021/gnn-intro.

\bibitem{goyal2022graph}
Naman Goyal and David Steiner,
\newblock ``Graph neural networks for image classification and reinforcement
  learning using graph representations,'' 2022.

\bibitem{Zhang_2022}
Bingyi Zhang, Rajgopal Kannan, Viktor Prasanna, and Carl Busart,
\newblock ``Accurate, low-latency, efficient sar automatic target recognition
  on fpga,''
\newblock in {\em 2022 32nd International Conference on Field-Programmable
  Logic and Applications (FPL)}. Aug. 2022, IEEE.

\bibitem{han2022vision}
Kai Han, Yunhe Wang, Jianyuan Guo, Yehui Tang, and Enhua Wu,
\newblock ``Vision gnn: An image is worth graph of nodes,''
\newblock in {\em NeurIPS}, 2022.

\bibitem{minibatch}
Arnab~Kumar Mondal, Vineet Jain, and Kaleem Siddiqi,
\newblock ``Mini-batch graphs for robust image classification,'' 2021.

\bibitem{cheng2021ba2m}
Qishang Cheng, Hongliang Li, Qingbo Wu, and King~Ngi Ngan,
\newblock ``Ba2m: A batch aware attention module for image classification,''
  2021.

\bibitem{hu2021on}
Yaochen Hu, Amit Levi, Ishaan Kumar, Yingxue Zhang, and Mark Coates,
\newblock ``On batch-size selection for stochastic training for graph neural
  networks,'' 2021.

\bibitem{yu2019identity}
Shuzhi Yu and Carlo Tomasi,
\newblock ``Identity connections in residual nets improve noise stability,''
  2019.

\bibitem{chest}
Daniel Kermany,
\newblock ``Labeled optical coherence tomography (oct) and chest x-ray images
  for classification,'' 2018.

\bibitem{gcnconv}
Thomas~N. Kipf and Max Welling,
\newblock ``Semi-supervised classification with graph convolutional networks,''
  2017.

\bibitem{tagconv}
Jian Du, Shanghang Zhang, Guanhang Wu, Jose M.~F. Moura, and Soummya Kar,
\newblock ``Topology adaptive graph convolutional networks,'' 2018.

\bibitem{sageconv}
William~L. Hamilton, Rex Ying, and Jure Leskovec,
\newblock ``Inductive representation learning on large graphs,'' 2018.

\bibitem{chebconv}
Michaël Defferrard, Xavier Bresson, and Pierre Vandergheynst,
\newblock ``Convolutional neural networks on graphs with fast localized
  spectral filtering,'' 2017.

\bibitem{bresson2018residual}
Xavier Bresson and Thomas Laurent,
\newblock ``Residual gated graph convnets,'' 2018.

\bibitem{vgg}
Karen Simonyan and Andrew Zisserman,
\newblock ``Very deep convolutional networks for large-scale image
  recognition,'' 2015.

\bibitem{vit}
Alexey Dosovitskiy, Lucas Beyer, Alexander Kolesnikov, Dirk Weissenborn,
  Xiaohua Zhai, Thomas Unterthiner, Mostafa Dehghani, Matthias Minderer, Georg
  Heigold, Sylvain Gelly, Jakob Uszkoreit, and Neil Houlsby,
\newblock ``An image is worth 16x16 words: Transformers for image recognition
  at scale,'' 2021.

\bibitem{smallvit}
Seung~Hoon Lee, Seunghyun Lee, and Byung~Cheol Song,
\newblock ``Vision transformer for small-size datasets,''
\newblock {\em CoRR}, vol. abs/2112.13492, 2021.

\bibitem{fastvit}
Pavan Kumar~Anasosalu Vasu, James Gabriel, Jeff Zhu, Oncel Tuzel, and Anurag
  Ranjan,
\newblock ``Fastvit: A fast hybrid vision transformer using structural
  reparameterization,'' 2023.

\bibitem{swin}
Ze~Liu, Yutong Lin, Yue Cao, Han Hu, Yixuan Wei, Zheng Zhang, Stephen Lin, and
  Baining Guo,
\newblock ``Swin transformer: Hierarchical vision transformer using shifted
  windows,'' 2021.

\bibitem{resnet}
Kaiming He, Xiangyu Zhang, Shaoqing Ren, and Jian Sun,
\newblock ``Deep residual learning for image recognition,''
\newblock in {\em 2016 IEEE Conference on Computer Vision and Pattern
  Recognition (CVPR)}, 2016, pp. 770--778.

\bibitem{alveoU200}
Xilinx,
\newblock ``Xilinx alveo u200 board,''
  \url{https://docs.xilinx.com/r/en-US/ds962-u200-u250/FPGA-Resource-Information}.

\bibitem{vitis}
``Vitis {HLS},''
  \url{https://www.xilinx.com/products/design-tools/vitis/vitis-hls.html}.

\end{thebibliography}

\end{document}